\documentclass[pdflatex,iicol]{sn-jnl}

\usepackage{graphicx}%
\usepackage{multirow}%
\usepackage{amsmath,amssymb,amsfonts}%
\usepackage{amsthm}%
\newcommand{\argmax}{\mathop{\rm arg~max}\limits}
\usepackage{mathrsfs}%
\usepackage[title]{appendix}%
\usepackage{xcolor}%
\usepackage{textcomp}%
\usepackage{manyfoot}%
\usepackage{booktabs}%
\usepackage{algorithm}%
\usepackage{algorithmicx}%
\usepackage{algpseudocode}%
\usepackage{listings}%
\usepackage[square, numbers]{natbib}

\theoremstyle{thmstyleone}%
\theoremstyle{thmstyletwo}%

\theoremstyle{thmstylethree}%

\begin{document}

\title[Article Title]{Model of Spatial Human-Agent Interaction with Consideration for Others}


\author*[1]{\fnm{Takafumi} \sur{Sakamoto}}\email{sakamoto.takafumi@shizuoka.ac.jp}

\author[1]{\fnm{Yugo} \sur{Takeuchi}}\email{takeuchi@inf.shizuoka.ac.jp}

\equalcont{These authors contributed equally to this work.}

\affil*[1]{\orgdiv{Education Development Center}, \orgname{Shizuoka University}, \orgaddress{\street{836, Ohya, Suruga-ku}, \city{Shizuoka}, \postcode{4228529}, \state{Shizuoka}, \country{Japan}}}

\affil*[2]{\orgdiv{Graduate School of Science and Technology, Shizuoka University}, \orgname{Shizuoka University}, \orgaddress{\street{3-5-1, Johoku, Naka-ku}, \city{Hamamatsu}, \postcode{4328011}, \state{Shizuoka}, \country{Japan}}}


\abstract{Communication robots often need to initiate conversations with people in public spaces. At the same time, such robots must not disturb pedestrians. To handle these two requirements, an agent needs to estimate the communication desires of others based on their behavior and then adjust its own communication activities accordingly. In this study, we construct a computational spatial interaction model that considers others. Consideration is expressed as a quantitative parameter: the amount of adjustment of one’s internal state to the estimated internal state of the other. To validate the model, we experimented with a human and a virtual robot interacting in a VR environment. The results show that when the participant moves to the target, a virtual robot with a low consideration value inhibits the participant’s movement, while a robot with a higher consideration value did not inhibit the participant’s movement. When the participant approached the robot, the robot also exhibited approaching behavior, regardless of the consideration value, thus decreasing the participant’s movement. These results appear to verify the proposed model’s ability to clarify interactions with consideration for others.}




\maketitle

\section{Introduction}
When a communication robot starts a conversation with a human in a public place, it needs to estimate the internal state of the human and adjust its behavior according to this estimated value. For example, people do not engage with each other without reason in public situations, and it is necessary to share a premise to start a conversation \cite{Goffman1963}.
When a robot needs to communicate with a human, it must estimate whether the human will accept an invitation to the conversation. Depending on the estimation result, the robot may decide to let the person pass without talking to them. When a person needs to converse with a robot, the robot must express its acceptance of the human’s initiation of conversation. If robots could perform such interactions with social consideration for others, they could make more progress in contributing to human society.

The issue of communication robots navigating public spaces is addressed within the research field of socially aware robot navigation. This study area extensively explores how robots navigate paths that either approach or avoid humans, with comprehensive research being conducted on both approaches (for reviews, see \cite{kruse2013human,rios2015proxemics,gao2021evaluation,mavrogiannis2023core}). Several studies have proposed methods that allow a robot to predict or classify human behavior in public situations and, based on its assessment, select an appropriate conversation partner \cite{kanda2009abstracting,satake2012robot}. Other research has focused on enhancing the predictability of robot movements, such as generating predictable avoidance trajectories \cite{mavrogiannis2018social} or investigating how a robot’s motor noise affects humans' acceptance of its approach \cite{joosse2021making}. By applying the findings of these studies, robots are enabled to exhibit spatial behaviors that are perceived as comfortable by humans, with each study developing a behavior model for robots that is adaptable to a specific scenario or scene. Conventionally, the tasks of approaching and avoiding humans are typically treated as distinct scenarios within these investigations.

Communication robots must determine their objectives based on their role, the situation, and the state of the people around them, necessitating a model that can be applied to multiple scenarios. This involves estimating the internal states of humans, such as desires and intentions. In response to this requirement, extensive research has been devoted to estimating people’s engagement with robots (for a review, see \cite{salam2023automatic}). For instance, studies have shown that robots can sense multimodal information to estimate human engagement, detecting the initiation of engagement in scenarios where individuals who intend to interact with the robot coexist with those who do not \cite{vaufreydaz2016starting}.

Moreover, robots need to express engagement through their actions, which essentially means enabling humans to estimate the robot’s internal state. Research has been conducted on both aspects of estimation and expression by robots (for a review, see \cite{oertel2020engagement}). However, for robots to achieve sociable capabilities, as proposed previously \cite{breazeal2003toward}, where they possess internal states and the ability to adjust these states mutually, it is necessary to integrate estimation and expression. This integration allows for a nuanced interaction between robots and humans, where understanding and responding to the complex dynamics of human engagement become possible.

In social interactions, humans and robots engage in a mutual estimation and adaptation of each other’s internal states, highlighting the necessity for bidirectional information flow: from humans to robots, which involves the perception and estimation of human states, and from robots to humans, which involves making the robot’s state understandable to people. Models such as those of Maniscalco et al. (2022) addressed a robot’s capability to perceive human states and express its own state in an understandable manner \cite{maniscalco2022bidirectional}. Similarly, Tanevska et al. (2020) showed that robots adjust their internal states in response to human stimuli, altering their engagement expression \cite{tanevska2020socially}. These studies are promising for designing sociable robots that can express socially adept behavior choices based on the dynamics of their internal states. However, when robots must choose actions based on both their and the humans’ internal states, they face a trade-off between prioritizing their own desires (the tasks they are engaged in) and those of the humans they encounter. Currently, no model presupposes a shared internal state between robots and humans, thus not addressing the trade-off between robot and human desires in their interaction dynamics.

Assuming a shared internal state between humans and robots might allow for the expression of robot sociability through controlling the dynamics of these internal state values. Prior research has proposed models of approach-avoidance behaviors and internal states at the initiation of communication, and their viability was validated \cite{sakamoto2018simulation,sakamoto2021investigation}. One key feature of such a model is its expression of the desire for engagement from oneself toward another and vice versa, within a range of values from -1 to 1. Based on these values, the model generates approach behaviors toward individuals with whom engagement is desired and avoidance behaviors toward those with whom it is not. However, despite the internal states being represented in a computable form, the dynamics of changes in these internal states were not addressed by this type of model. To expand the model to social robots, it is necessary to add the capability to estimate others’ desires based on behavior and then to adjust one’s desires based on these estimated values. This augmentation would permit the development of robots capable of adjusting their behavior not only based on specific roles or scenarios but also in response to the individuals they interact with, thus enhancing the adaptability and relevance of robot actions in social contexts.

This study aims to model spatial interactions that incorporate consideration for others at the initiation of communication. Initially, we extend the cognitive model from prior research, which generates approach and avoidance behaviors based on the internal states of individuals and their dialogue partners. Specifically, functions for estimating internal states and for the dynamics of changes in these states are introduced. This framework allows consideration to be defined as the amount by which an individual’s internal state changes in response to another’s internal state.

Next, through computer simulations, the study examines the alterations in generated trajectories relative to the level of consideration and implements these findings in virtual robots within a VR environment. Experiments observing interactions between these virtual robots and humans are conducted to analyze how the consideration parameter values affect participants’ behaviors. This investigation was made to confirm whether, in scenarios where robots are designed to engage with humans, virtual robots programmed with high levels of consideration indeed avoid impeding others and seek engagement only when necessary, whereas virtual robots with low levels of consideration persistently approach those who do not wish to interact. Consequently, this approach assesses the capability of the proposed model to accurately reflect human-robot interactions that include elements of consideration.

\section{Model of interaction with social consideration for others}
\begin{figure*}[tb]
	\centering
		\includegraphics[width=\linewidth]{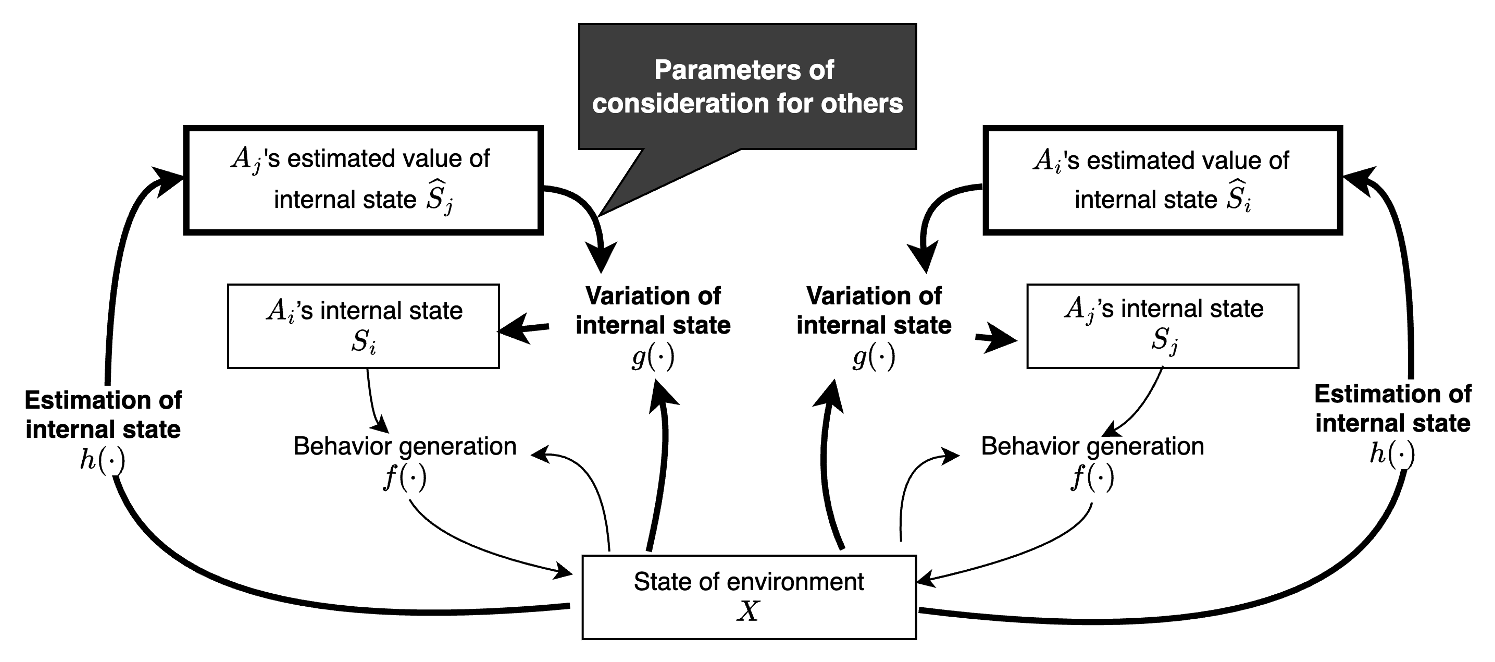}
		\caption{Functions and variables required to describe interactions with consideration for others}\label{fig:model_diff}
\end{figure*}

Generally, considering others involves a mental state and attitude of supporting them as much as possible and respecting their intentions and actions. On the other hand, the cognitive aspect of interaction can be defined as adjusting one’s internal condition in response to the internal condition of others (especially their desires) and an estimated value of their behavior. Therefore, to quantitatively describe interactions between agents (people) that involve consideration for others, we must define variables and functions as shown in Fig. \ref{fig:model_diff}.
In particular, the behavior of an agent that exercises consideration for others can be expressed through a function that estimates the value of another’s internal condition and a function that adjusts the relative value of its internal condition based on that estimated value (bold line in Fig. \ref{fig:model_diff}).

Based on this processing framework, we adopt a model that generates the behavior of considering another person at the beginning of an interaction through a function that minimizes the difference between the estimated level of that person’s desire and one’s own level of desire. At the same time, this model expresses the level of consideration for others through a quantitative parameter that expresses how little this difference declines.

A model of interaction based on approach and avoidance behaviors in a communication-initiation scene between two agents is outlined below. In describing the internal conditions, the action generations, and the estimation of the internal conditions, only a summary is given when there is overlap with the previous research \cite{sakamoto2018simulation}.

\subsection{Internal Condition and Behavior Generation}
We use the model of a previous study \cite{sakamoto2018simulation} for the action-generation function of agents according to the value of the assumed internal condition. In this model, the physical interaction between agents $A_{1}$ and $A_{2}$ is represented by the temporal changes in environment ${\bf x}_{12}=\{r_{12}, \theta_{12}, \theta_{21}\}$.
Note that $r_{12}$ represents the distance between $A_{1}$ and $A_{2}$ and that $\theta_{12}$ and $\theta_{21}$ represent the absolute values of the relative angle from $A_{1}$ and $A_{2}$, respectively.
An agent’s internal condition represents variables that facilitate or inhibit its behavior. Here, we address two variables of the internal condition in the communication-starting scene: the preference (Control) for involvement from oneself to another agent and the preference (Acceptance) for involvement from the other agent to oneself. The internal condition of ${\bf A}_{1}$ with respect to ${\bf A}_{2}$ is denoted by ${\bf s}_{1\to 2}=(c_{1}, a_{1})$ and the internal condition of ${\bf A}_{2}$ by ${\bf s}_{2\to 1}=(c_{2}, a_{2})$.
${\bf s}_{1\to 2}$ represents the value of the internal condition of $A_{1}$, and ${\bf s}_{1\to 2}=(c_1,a_1)\in[-1,1]^2$.
Each action of an agent is represented by a temporal change in ${\bf x}_{12}$.
The action of $A_{1}$, $\Delta_{1} {\bf x}_{12}$, is represented by the function $f({\bf x},{\bf s})$ of behavior generation.

\begin{figure}[htb]  
	\centering
	\includegraphics[width=\linewidth]{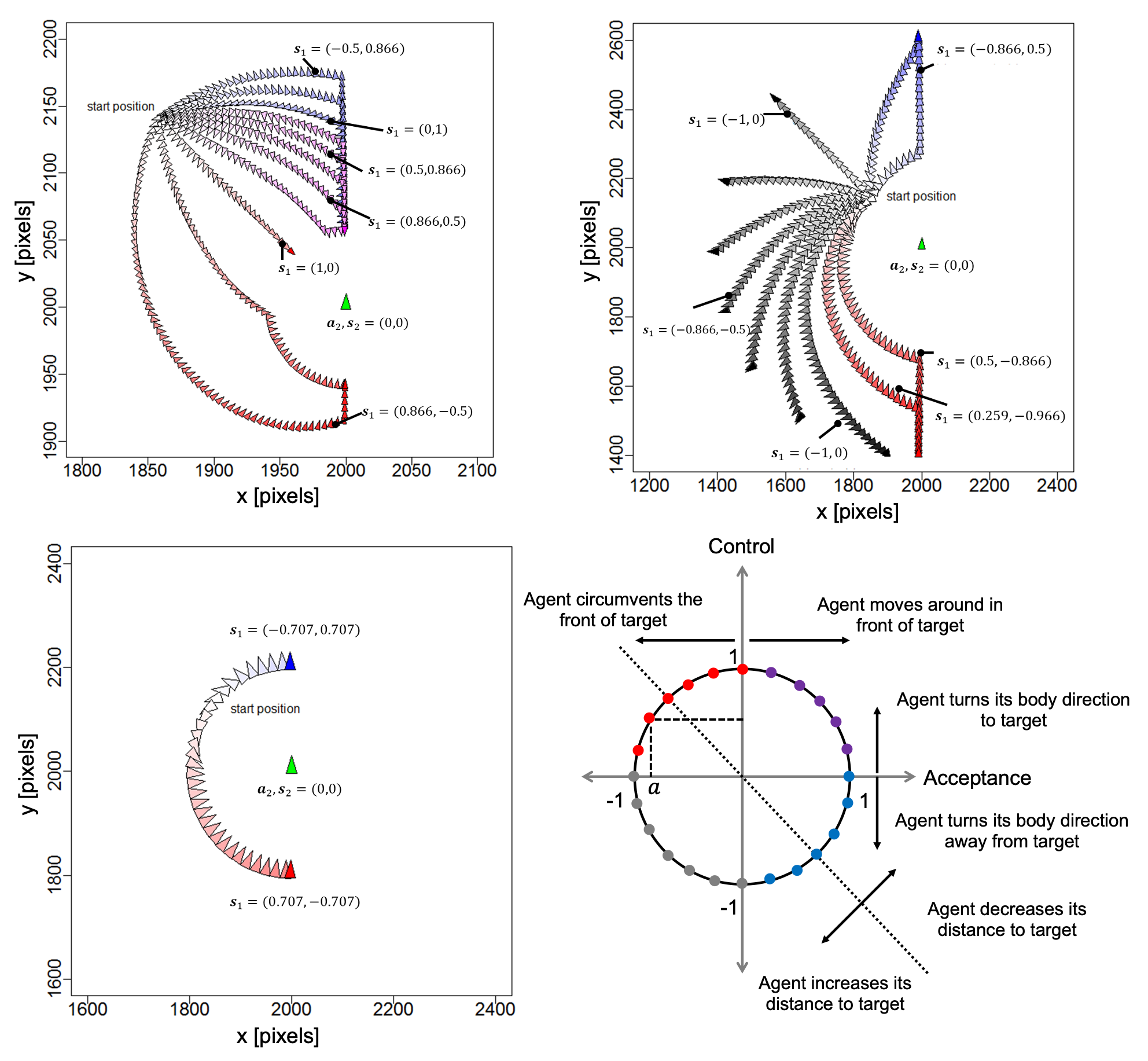}
	\caption{Examples of approach and avoidance actions generated according to internal state values \cite{sakamoto2018simulation}}\label{fig:sim_behavior}
\end{figure}

Based on these variables, Fig. \ref{fig:sim_behavior} shows the approach and avoidance behaviors that are only generated when the internal condition and function, $f$, for behavior generation are specified.

\subsection{Estimation Function of Internal Condition}
In generating behaviors that incorporate consideration, the internal conditions of other agents must be estimated. Here, they are estimated by applying the function of behavior generation.

When $A_{1}$ infers the internal condition of $A_2$ from behavior $\Delta_2{\bf x}_{12}$, ${\bf x}_{12}$ is an observable variable.
${\bf \phi}_{2}$ must be inferred, but since it is a variable that represents behavioral characteristics, we assume that it can be roughly inferred by observing the behavior. Then the estimated value of ${\bf s}_{2\to 1}$ by $A_{1}$, $\hat{\bf s}_{2\to 1}$, can be expressed by the following equation using function $h$:
\begin{eqnarray}
	\hat{\bf s}_{2\to1} = h\left( x_{12}, \Delta_{2}{\bf x}_{12} \right) \nonumber = \argmax_{\hat{\bf s}}\left( L\left(\Delta_2{\bf x}_{12} \right)\right).\label{eq:def_h}
\end{eqnarray}
However, $L(\cdot)$ is defined as follows:
\begin{equation}
	L(\cdot)=1-l\left(f({\bf x},{\bf \hat{s}})- \Delta_2{\bf x}_{12}\right).
\end{equation}
Here, $l$ represents a function that normalizes the difference between the behaviors that can be generated by $f$ and $\Delta_2{\bf x}_{12}$, where $0\leq l(\cdot) \leq 1$.
This definition estimates the internal condition that can generate the action that is most similar to $\Delta_2{\bf x}_{12}$.
In this case, since ${\bf s}_{2\to 1}$ is a two-dimensional variable, the solution can be approximated by a grid search.

\subsection{Function Representing the Change in Internal Condition}
The change in the internal condition is expressed with function $g$.
When dealing only with changes in the internal condition based on behavioral outcomes, the change in the internal condition of $A_{1}$, $\Delta {\bf s}_{1\to 2}$, is represented by the following equation:
\begin{equation}
	\Delta {\bf s}_{1\to 2} = g({\bf x}_{12}, {\bf s}_{1\to 2}; {\bf
	 \psi}_1).
\end{equation}
Here, ${\bf \psi}_1$ represents such cognitive properties as the speed of the change of the internal condition of $A_{1}$.
As in our previous study, to represent the internal state’s change when $A_{1}$ is concerned with $A_{2}$, we extend function $g$:
\begin{equation}
	\Delta {\bf s}_{1\to 2} = g({\bf x}_{12}, {\bf s}_{1\to 2}.
	 \hat{\bf s}_{2\to 1}; {\bf \psi}_1)\label{eq:def_g}
\end{equation}
This procedure determines the level of consideration the agent will give to the other agent, depending on the values specified for function $g$ and parameter ${\bf\psi}$.

\subsection{Parameters for Consideration of Others}
In this study, consideration for others is represented by the value of a single parameter, $\psi$.
Since two preferences, which are internal condition variables, represent the differences in the direction of the preference for involvement from one agent to another, it follows that the value of Control in one agent corresponds to the value of Acceptance in the other. In other words, if the Control value of one agent and the Acceptance value of another agent are identical, their preferences will be satisfied. Therefore, we define function $g$ of the internal condition change to feedback the gap between the Control and Acceptance values. The change in the internal condition of $A_{1}$ is represented by the following equations:
\begin{equation}
	\Delta c_{1} = - \psi_1  (\hat{a}_2 - c_1),
\end{equation}
\begin{equation}
	\Delta a_{1} = - \psi_1  (\hat{c}_2 - a_1).
\end{equation}

In this case, parameter $\psi_{1}$ represents the amount by which $A_{1}$ changes its internal condition in consideration of $A_{2}$.
When $\psi_{1}$ is positive, $A_{1}$ exhibits behavior that is considerate of $A_{2}$.
The larger the value of $\psi_{1}$ is, the faster the adjustment of the internal condition of $A_{1}$.
On the other hand, there are cases where $\psi_{1}$ takes a negative value. If such a case occurs, the action will ignore the preferences of the other agent, which is the opposite behavior of consideration.

\section{Interaction Experiments in VR Environment}
In this section, we experimentally investigate the impact of considerate behavior generated by our model on the interaction between a human and an agent, as outlined in the previous section. From this experiment, we verify that our model can represent behaviors that exhibit consideration for others.
The experiment was conducted in a VR environment, where a virtual robot, which appears to move in the VR environment, is used as an agent to analyze the human responses to human-robot interactions.

\subsection{Experiment Environment}
\begin{figure}[tb]
	\centering
		\includegraphics[width=\linewidth]{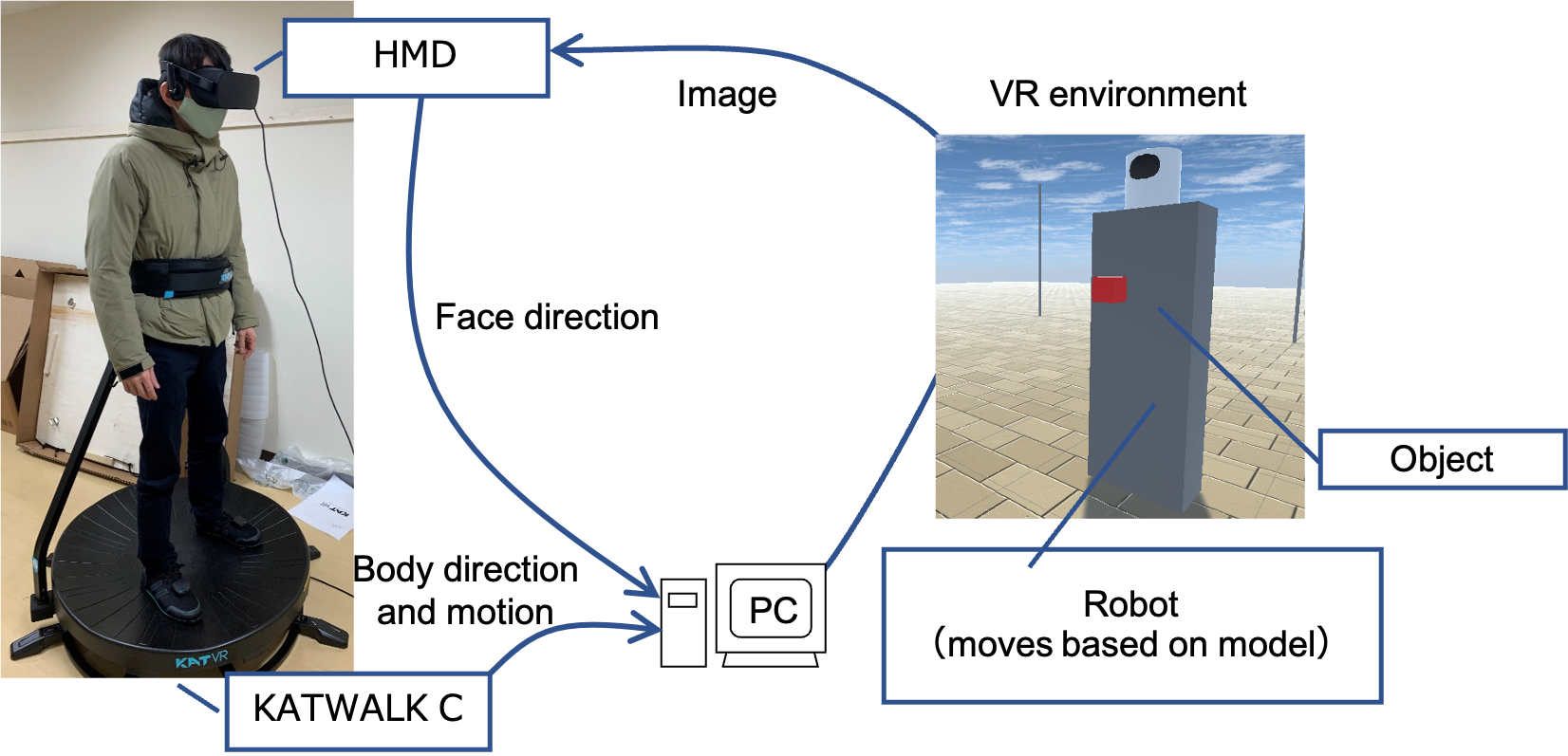}
		\caption{Experiment environment}\label{fig:ex_environment}
\end{figure}

Figure \ref{fig:ex_environment} shows an overview of our experiment’s environment, which is a VR environment built with Unity.
The participants wore a head-mounted display (HMD, Oculus Rift) to view the visual information in the VR environment.
They also wore shoes to which sensors were attached and moved and changed direction in the VR environment on a treadmill (KATWALK C).

We placed poles at six locations in the field of the VR environment and positioned the robot in the center. The poles functioned as the target positions for the experimental tasks. To reduce the information given by the virtual robot’s appearance, it was comprised of a combination of abstractly shaped objects. For example, its body part is a rectangle, its head is a cylinder, and part of its head is colored black to indicate its body direction (Fig. \ref{fig:ex_environment}). The robot generated behaviors based on the model described in the previous section according to the experimental conditions.

\subsection{Procedure}
\begin{figure*}[htb]
	\centering
		\includegraphics[width=32em]{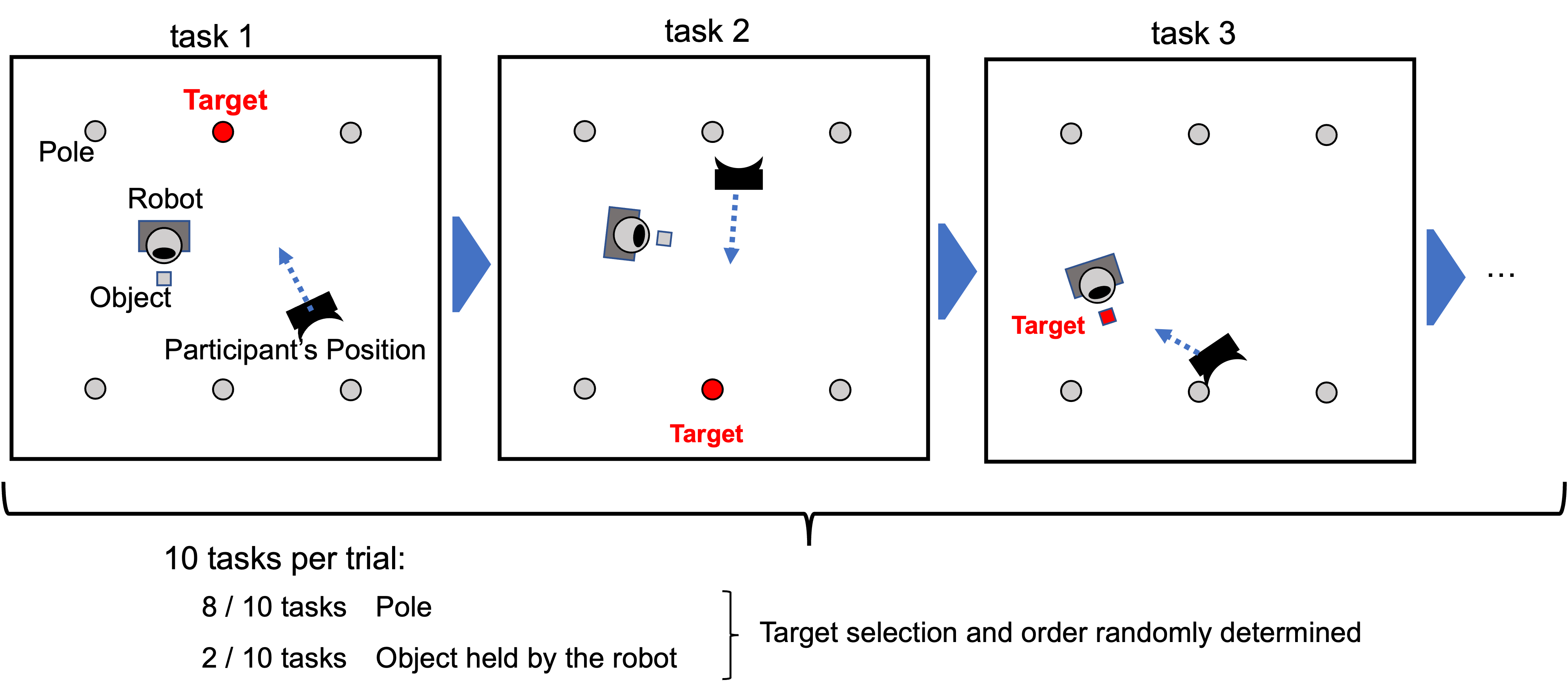}
		\caption{Experimental tasks}\label{fig:ex_tasks}
\end{figure*}

After the participants agreed to participate, we explained the experimental task to them. Figure \ref{fig:ex_tasks} shows the experimental task for one trial. Participants approached the target designated for each task, that is, either a pole or an object held by the virtual robot. The color of the pole or object changed to red when it was designated as the target. One task was completed when the participant reached the designated target, and participants performed the task 10 times during one trial. A pole was designated as the target 8 times and an object 2 times. The order in which targets were designated was random, but the object held by the robot was never designated consecutively. A pole was randomly assigned from the three poles on the side opposite to the previously assigned one. The robot’s initial position was in the center of the field, and it returned to this initial position for the start of each task. This positioning required the participant to pass close to the virtual robot.

We explained to our participants that “The robot cannot determine the target color and knows nothing about its given task.” We also told them that “The robot is currently handing off the objects it is holding.”

First, to get accustomed to using KATWALK C in the VR environment, the participants moved between poles several times while wearing the HMD. After they had become used to walking through the VR environment, the virtual robot was introduced but remained stationary as they practiced moving between poles several more times.

Subsequently, an interaction was initiated between a human and a virtual robot that operated based on the model described above. Participants performed one trial, i.e., ten approaches to the target. After removing their HMDs and stepping off KATWALK C, they completed a questionnaire.

Then, according to the situation, trials and questionnaires were conducted for a robot with different parameter values (maximum of three trials and questionnaires). We explained that our research ethics policy allowed them to stop participating in the experiment at any time. 

\subsection{Experimental Conditions}
The experiment was conducted under four conditions: the virtual robot had social consideration $\psi_{\rm Robot}$ values of 0.001, 0.005, and 0.01, and, as the control condition, it randomly walked (RW).

\begin{figure*}[tb]
\centering
\includegraphics[width=\linewidth]{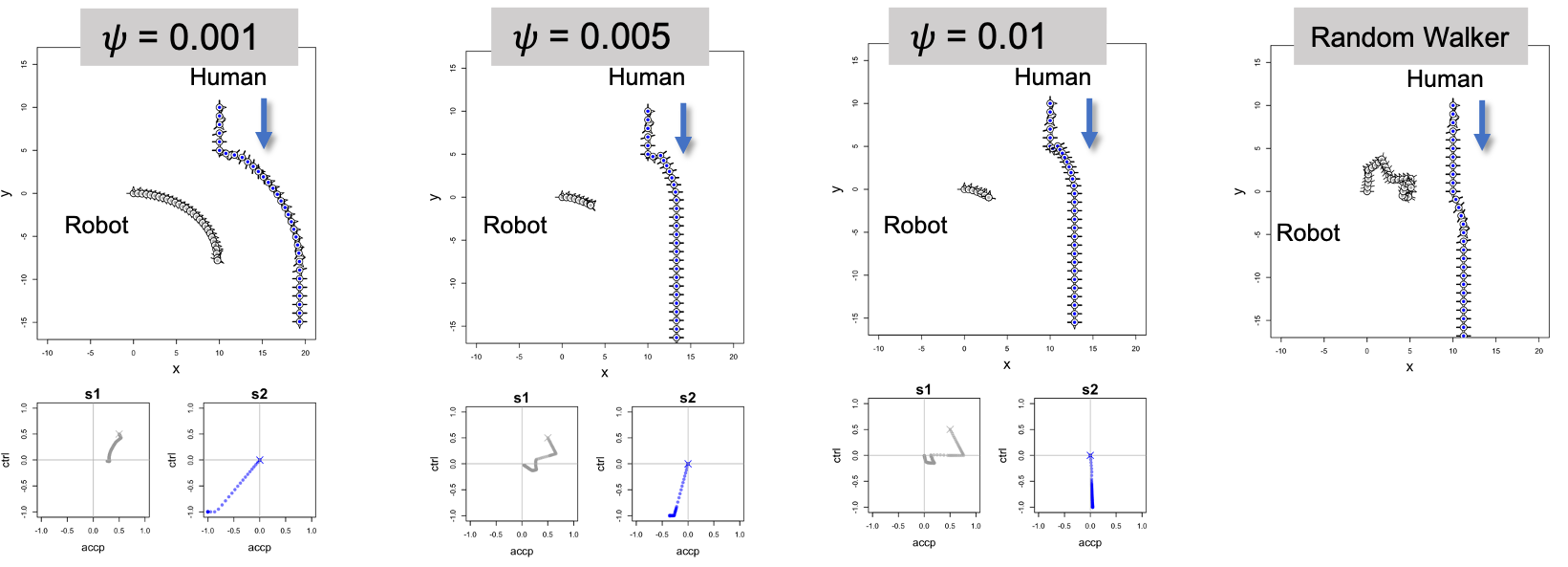}
\caption{Simulated trajectories when participants reject the robot in each condition}\label{fig:ex_con_sim2}
\end{figure*}

\begin{figure*}[tb]
\centering
\includegraphics[width=\linewidth]{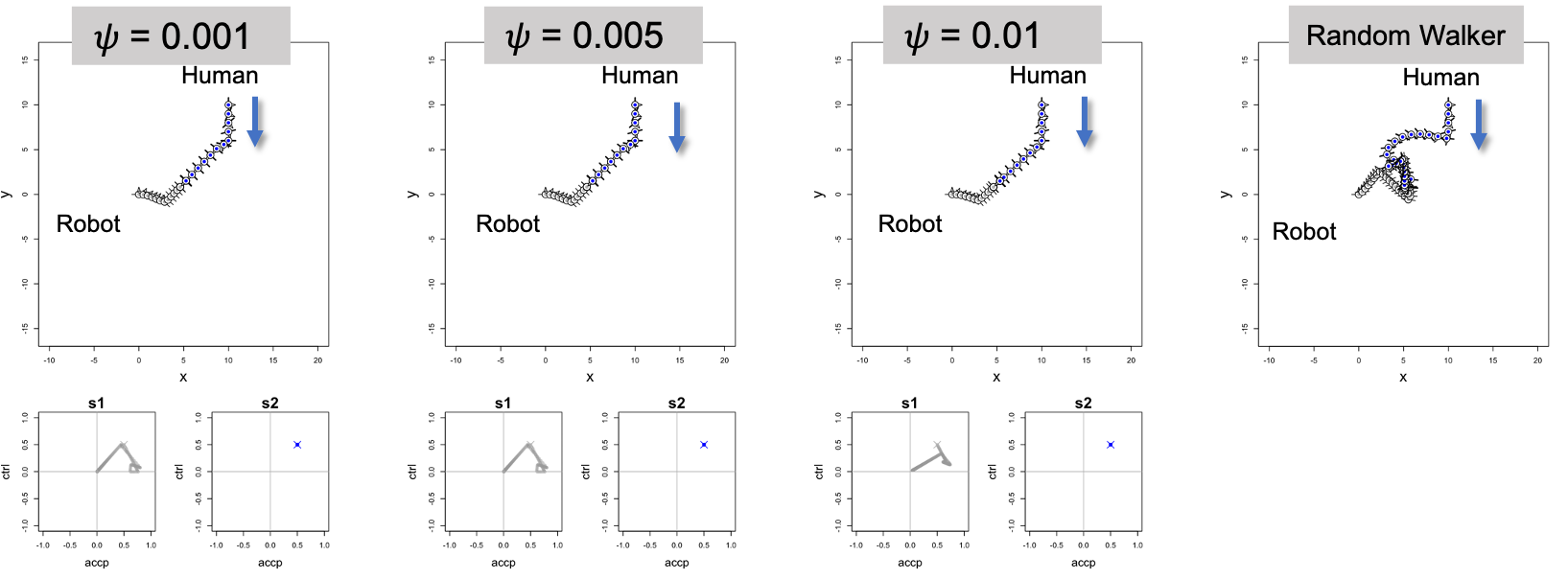}
\caption{Simulated trajectories when participants approach the robot in each condition}\label{fig:ex_con_sim3}
\end{figure*}

Figures \ref{fig:ex_con_sim2} and \ref{fig:ex_con_sim3} show simulated virtual robot behavior for each condition. Note that the upper part of these figures show the changes in the position and body direction of the robot and the hypothetical participant, while the lower part shows the changes in the internal state. The trajectories in Fig. \ref{fig:ex_con_sim2} and \ref{fig:ex_con_sim3} correspond to $\psi_{\rm Robot}$ values of 0.001, 0.005, and 0.01 from the left, and the rightmost one is a randomly walking robot. Since the robot was configured to distribute objects to the participants, the initial value of its internal condition was set to $\left(c_R^{(0)}, a_R^{(0)}\right)=\left(0.5, 0.5\right)$.
%
%


A rejective response to the robot’s approach can be simulated by setting a negative $\psi$ value. Figure \ref{fig:ex_con_sim2} shows the trajectories of the robot and the participant when the participant reacts negatively to the approaching robot. Comparing the trajectories for each value of $\psi_{\rm Robot}$, the larger it is, the more quickly the robot interrupts its approach to the hypothetical participant, and the smaller the amount is, the larger the curve of the participants’ trajectories.

Figure \ref{fig:ex_con_sim3} shows the simulated trajectory of a hypothetical participant approaching an object held by the robot. The internal state of the hypothetical participant is set to 0.5 to approach the robot. The trajectories do not differ according to the value of $\psi_{\rm Robot}$, since the preferences of the robot and the hypothetical participant are identical. On the other hand, the hypothetical participants need to move extensively to get closer to the random walker, as shown in the figure on the far right.


\subsection{Working Hypotheses}
Based on the results of the simulations described above, this study tests the following three working hypotheses:

\begin{description}
	\item[H1:] The behavior of robots with a lower consideration value $\psi_{\rm Robot}$ more greatly inhibits the movement of individuals not needing to approach the robot than the behavior of robots with a higher consideration value.
	\item[H2:] Repeated interactions with robots having a lower consideration value $\psi_{\rm Robot}$ will increase the tendency of individuals to avoid those robots.
	\item[H3:] Robots following the model will reduce the movement of individuals needing to approach the robot, regardless of the robots’ consideration value, compared to Random Walkers.
\end{description}

When the consideration value of a robot is low, even if the robot’s perceived human value of desire toward interaction is low, the robot continues to approach individuals due to the smaller amount of change in its internal state per unit time. Therefore, it is assumed that the behavior of individuals trying to avoid the robot will be more pronounced when a pole is designated as the target (H1).

Furthermore, as individuals repeatedly interact with the robot, they are expected to adjust their behavior based on predictions about the robot’s actions and their perception of the robot. This adjustment is expected to lead to more pronounced avoidance behavior toward robots with lower consideration values (H2).

When the desire of individuals toward a robot is high and they approach the robot, robots operating based on the model will estimate this state of desire and continue to approach the individuals. Therefore, it is believed that in the three conditions where robots operate based on the model, the movement of the experiment’s participants will decrease when the object possessed by the robot is designated as the target (H3).

By testing these hypotheses with actual experimental data, this study aims to evaluate the validity of the proposed model as a tool for representing consideration.

\subsection{Measurement and Analysis}
As behavioral data, the coordinates of the participant and the robot in the virtual environment were recorded at 50-ms cycles; KATWLAK C sensed the participant’s body direction and the HMD sensed the participant’s face direction. To test the working hypotheses, two indices were used based on the behavioral data: the gap between the participant’s direction of movement (cf. path irregularity \cite{guzzi2013human}) and the target pole position and, on the other hand, the participant’s amount of movement when the target was the object held by the robot.

\begin{figure}[tb]
\centering
\includegraphics[width=20em]{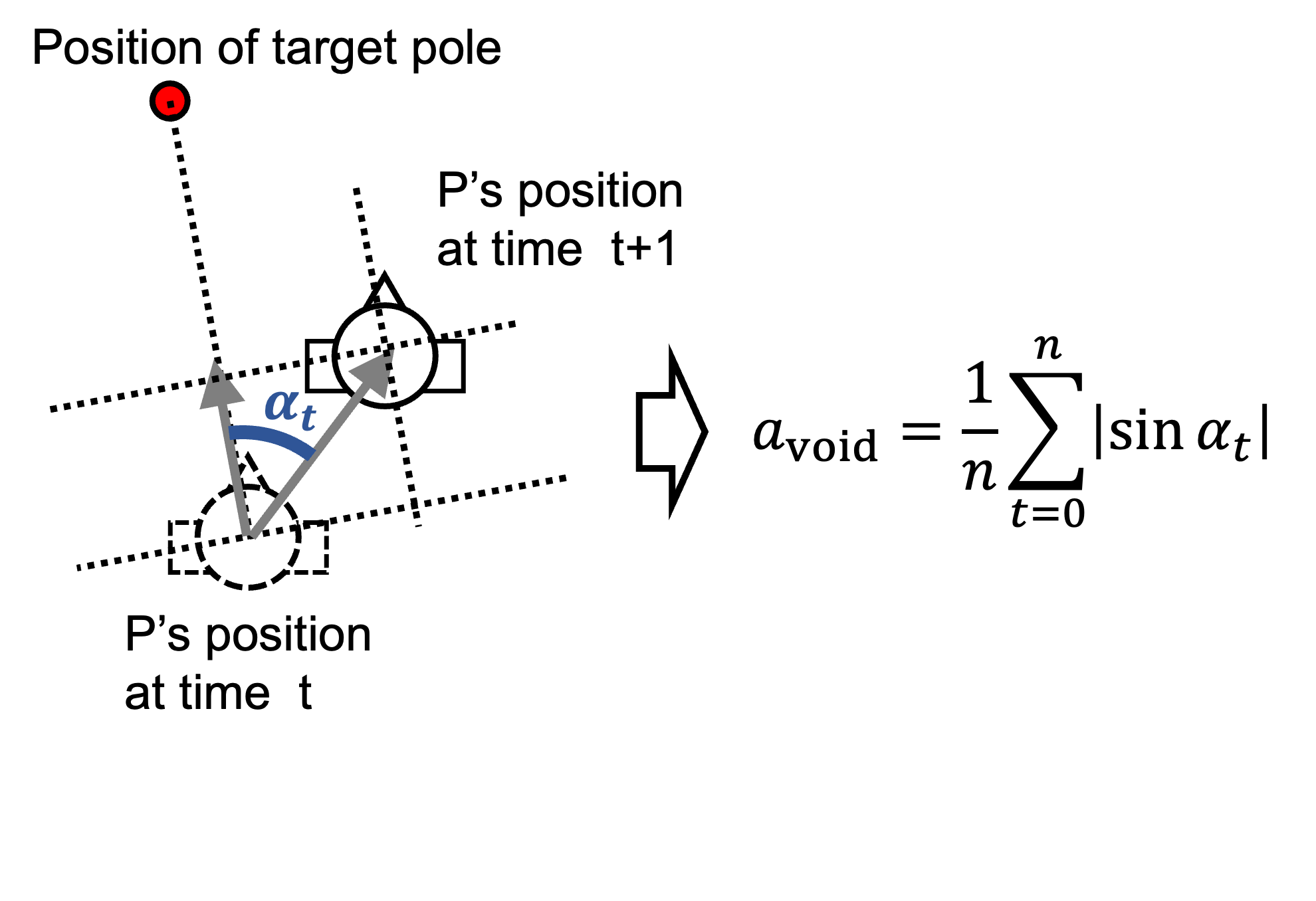}
\caption{Vertical component of movement vector used to assess amount of virtual robot’s avoidance}\label{fig:ana_avoid}
\end{figure}

Figure \ref{fig:ana_avoid} shows how to calculate the gap between the target pole and the participant’s direction of movement. The gap of the angle at time $t$ is $a_t$, and the average value of the gap in one movement task is the amount of the robot’s avoidance $a_{\rm void}$. Here, $a_{\rm void}$ is calculated from the following equation:
\begin{equation}
  a_{\rm void} = \frac{1}{n}\sum_{t=0}^{n}|\sin{\alpha_t}|,
\end{equation}
where $n$ is the number of frames until the participant reaches the target (one frame is logged every 50 ms). The amount of avoidance, $a_{\rm void}$, was determined in a task where the pole was designated as the target. This indicator was used to validate H1. The $a_{\rm void}$ values in the trial--i.e., the average of the $a_{\rm void}$ values for the eight tasks--were used for the one-way between-group ANOVA.

For the validation of hypothesis H2, it is necessary to demonstrate that $a_{\rm void}$ increases with the number of movements during each trial. To do this, the relationship between $a_{\rm void}$ and the number of movements is represented through linear regression, and the distribution of regression coefficients for each condition is estimated using Bayesian statistical methods. The Bayesian model used for estimation is expressed by the following equations:

\begin{align}
	\beta_{0,i} &\sim \text{Normal}(\mu_{0,c}, \sigma_{0}) ,\\
	\beta_{1,i} &\sim \text{Normal}(\mu_{1,c}, \sigma_{1}) ,\\
	a_{i,m} &\sim \text{Normal}(\beta_{1,i}m + \beta_{0,i}, \sigma_{2}).
\end{align}

Here, $i$ represents each trial, $m$ represents the number of movements made within a trial, and $c$ represents each condition. $a_{i,m}$ denotes the $a_{\rm void}$ value for the $m$th movement in trial $i$, which is represented by a linear regression model with $\beta_{1,i}$ as the slope and $\beta_{0,i}$ as the intercept. In other words, $\beta_{1,i}$ represents the amount of increase in $a_{\rm void}$ during the $i$th trial. $\beta_{1,i}$ and $\beta_{0,i}$ are assumed to be values generated based on a normal distribution, with their mean values varying according to the corresponding experimental condition $c$. Therefore, this Bayesian model assumes that the regression coefficient and intercept for trials under a certain condition are generated based on a normal distribution and that the amount by which participants avoid the robot is determined according to these values. Bayesian statistics allow for the estimation of the distributions of $\beta_{1}$ and $\beta_{0}$ for each condition. Specifically, if hypothesis H2 is valid, it is expected that the values of $\beta_{1}$ under conditions with lower consideration values would be positive.

In this study, MCMC was performed using RStudio and Rstan (Ver. 2.21.8) to estimate the distributions of these coefficients. The number of chains was set to 4, with a chain length of 10,000 and a burn-in period of 5,000, resulting in a total of 20,000 samples for the estimation of the posterior distribution.


For the validation of hypothesis H3, we compared the movements of experiment participants required to reach the target held by the robot. If the robot could estimate the intentions of the participants and approach them, then the movements of the participants would decrease. Therefore, the movement of the participant was calculated in each trial where the target was the object held by the robot, and thus the average value was obtained. From this value, an analysis of variance (ANOVA) was used to perform comparisons between conditions.

Our participants were 23 undergraduate and graduate students (mean age=18.9, standard deviation of age=4.06, 16 males and 7 females). Participants completed one to three trials, depending on their situation. We collected data for 54 trials: 11 trials for the 0.001 condition; 16 trials for the 0.005 condition; 14 trials for the 0.01 condition; and 13 trials for the random walker condition.

\subsection{Results and Discussion}
\subsubsection{Interaction when a pole was the target}
\begin{figure*}[tb]
	\centering
		\includegraphics[width=\linewidth]{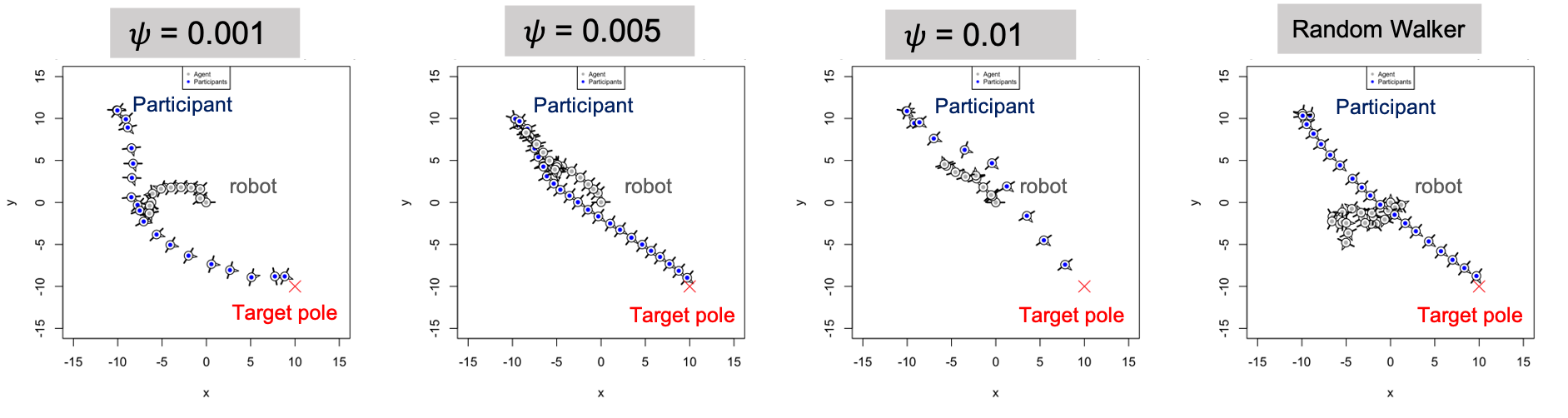}
		\caption{Example trajectories of a participant and a robot in a task where a pole is specified as target}\label{fig:res_trj_avo}
\end{figure*}

Figure \ref{fig:res_trj_avo} shows an example of the observed interaction between the participant and the virtual robot when a pole was designated as the target in each experimental condition. Figure \ref{fig:res_trj_avo} shows the trajectory of the scene where the participant moved from the upper-left position to the lower-right pole as a task. The leftmost trajectory ($\psi_{\rm Robot}=0.001$  condition) is clearly more curved than the other trajectories.

\begin{figure}[tb]
	\centering
		\includegraphics[width=\linewidth]{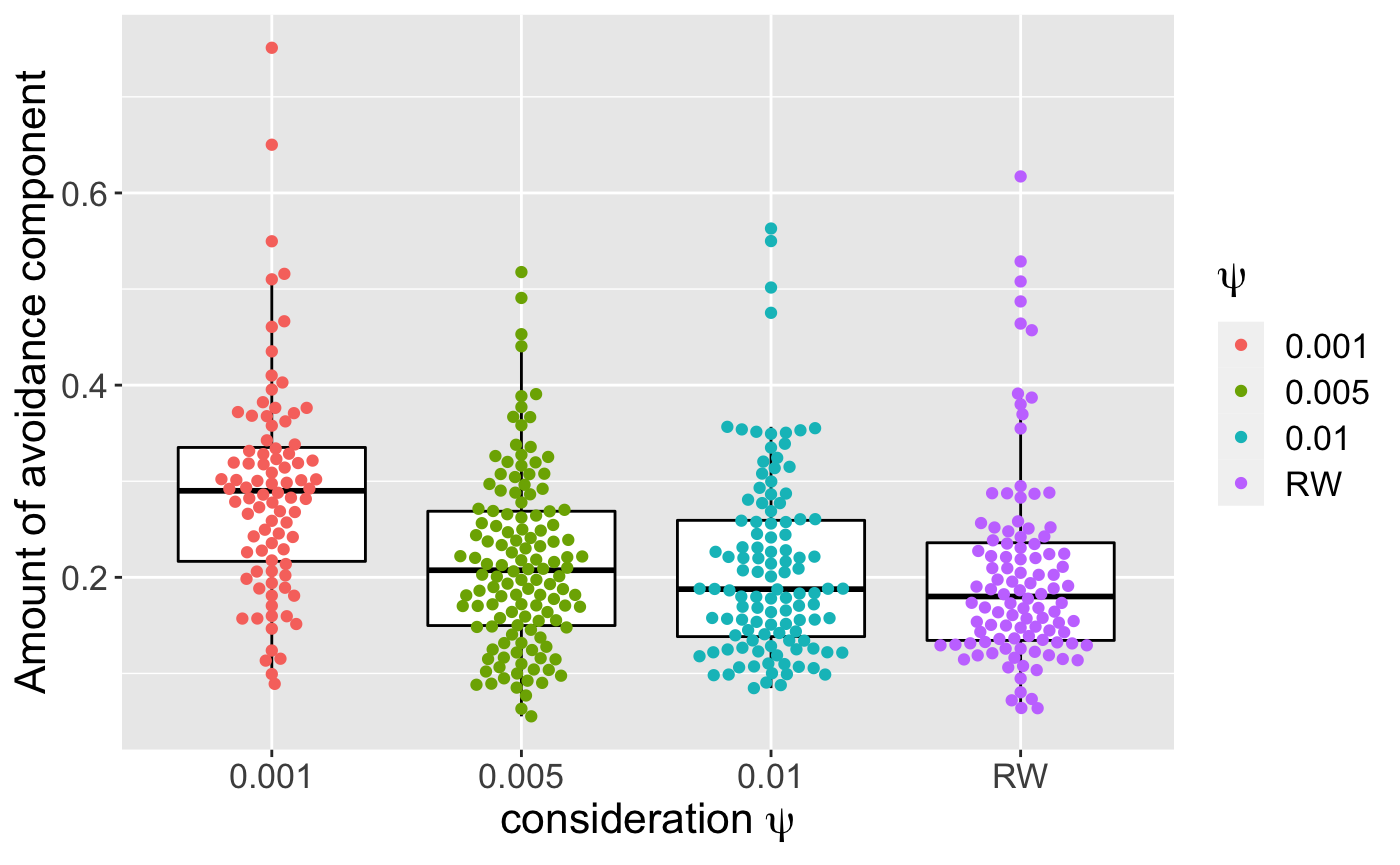}
		\caption{Comparison of $a_{\rm void}$ under each condition}\label{fig:res_avo}
\end{figure}

Figure \ref{fig:res_avo} shows the $a_{\rm void}$ values of the participants in trials under each condition. The value of $a_{\rm void}$ was larger under the condition of $\psi_{\rm Robot}=0.001$ at the left end of \ref{fig:res_avo} ($M=0.29, SD=0.04$).
On the other hand, only small differences were observed in the $\psi_{\rm Robot}=0.005$, $\psi_{\rm Robot}=0.01$ and RW conditions ($M=0.22$, $SD=0.06$; $M=0.21$, $SD=0.06$, $M=0.20$, $SD=0.08$, respectively). One-way between-group ANOVA results show significant differences in means of $a_{\rm void}$ among the four conditions ($F(3,50)=5.08$, $f=0.55$, $p < 0.01$). The results of multiple comparisons by HSD show that the $\psi_{\rm Robot}=0.001$ condition had significantly higher $a_{\rm void}$ mean values than the other three conditions. The $a_{\rm void}$ differences among the $\psi_{\rm Robot}=0.005$, $\psi_{\rm Robot}=0.01$, and RW conditions were not significantly different for any of the combinations compared.
%
%
%
%
The analysis shows that the gap between the participant’s direction of movement and the pole direction was significantly larger in the condition of robot consideration parameter $\psi_{\rm Robot}=0.001$ than in the other conditions. This result supports H1.

\begin{figure*}[tbp]
	\centering
	\includegraphics[width=\linewidth]{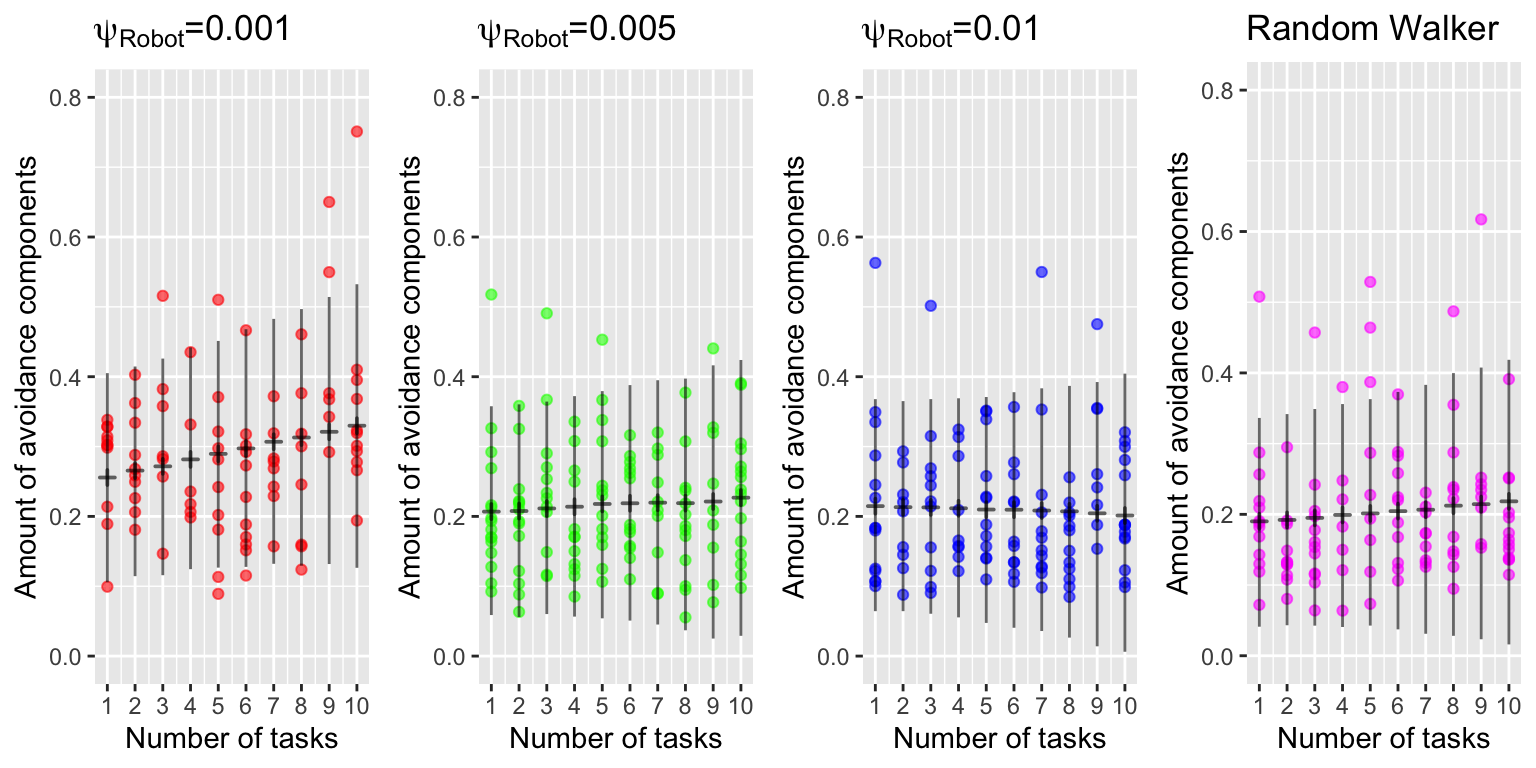}
	\caption{Comparison of participant movements when approaching the robot under each condition \label{fig:res_avo_trend}}
\end{figure*}

\begin{figure}[tbp]
	\centering
	\includegraphics[width=\linewidth]{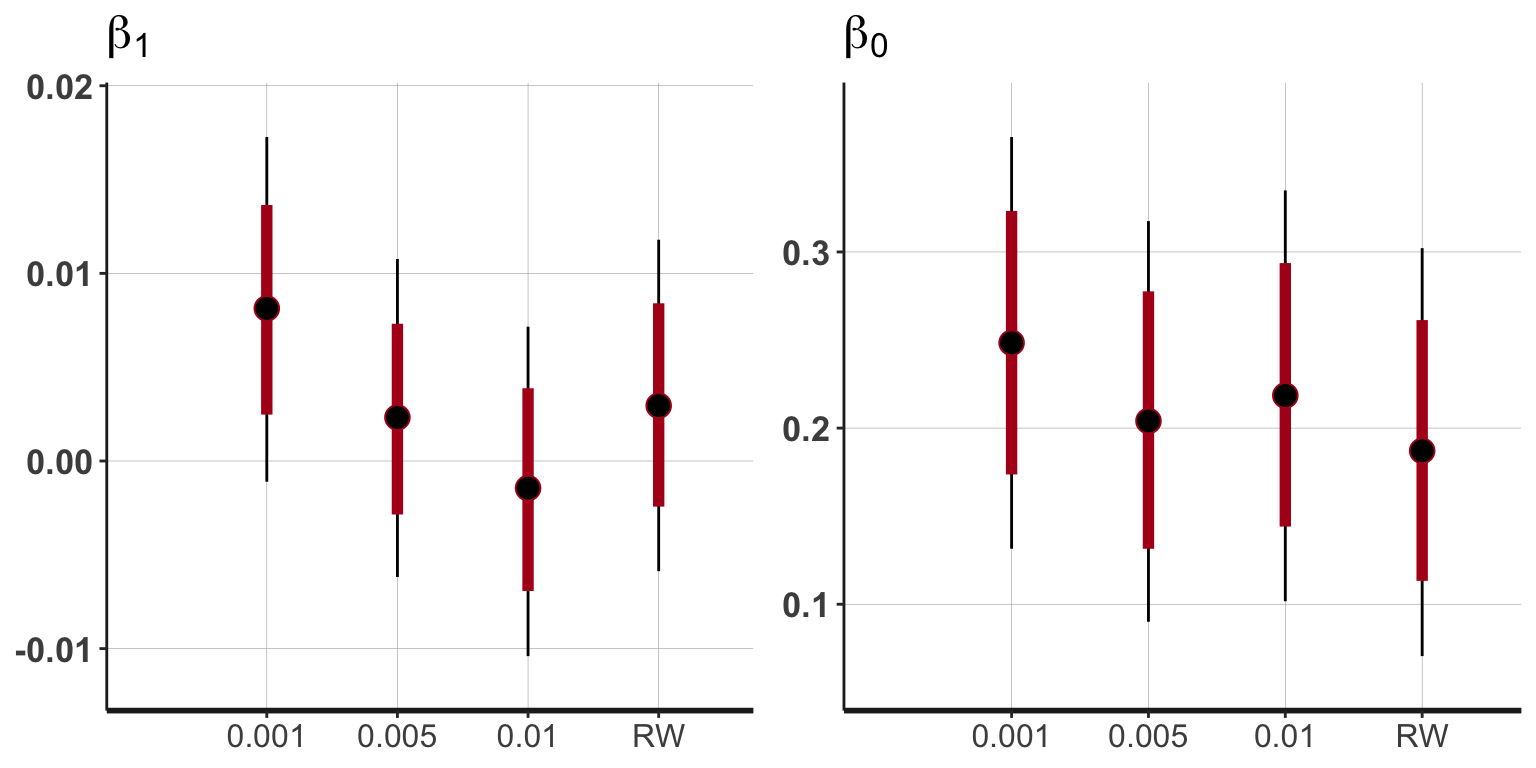}
	\caption{Distributions of $\beta_{1}$ and $\beta_{0}$ showing a comparison of participant movements when approaching the robot under each condition \label{fig:res_beta}}
\end{figure}

Figure \ref{fig:res_avo_trend} shows the avoidance parameter $a_{void}$ for each movement within the trials. Each point represents the $a_{void}$ obtained from actual experiments, and the gray dash mark indicates the mean and 95\% interval of the $a_{void}$ distribution estimated by MCMC. Under the $\psi_{\rm Robot}=0.001$ condition, there is a tendency for $a_{void}$ to increase from the first to the tenth movement within a trial. However, this increasing trend is not as evident in the other three conditions. For each condition, the distributions of $\beta_{1}$, representing this increasing trend, and $\beta_{0}$, representing the intercept, are shown in Figure \ref{fig:res_beta}.

In Figure \ref{fig:res_beta}, the black points represent the mean values, the thick red line represents the 80\% interval, and the black line represents the 95\% interval. The probability of $\beta_{1}>0$ estimated from the MCMC resampling results is 96.2\% under condition $\psi_{\rm Robot}=0.001$, 71.9\% for $\psi_{\rm Robot}=0.005$, 36.8\% for $\psi_{\rm Robot}=0.01$, and 76.9\% for the RW condition. This indicates that as the number of movements increases, the amount by which participants avoid the robot increases under condition $\psi_{\rm Robot}=0.001$. Additionally, the results show that as the consideration parameter increases, the probability of participants avoiding the robot decreases. This result supports H2.

\subsubsection{Interaction when an object held by the robot was the target}
\begin{figure*}[tb]
	\centering
		\includegraphics[width=\linewidth]{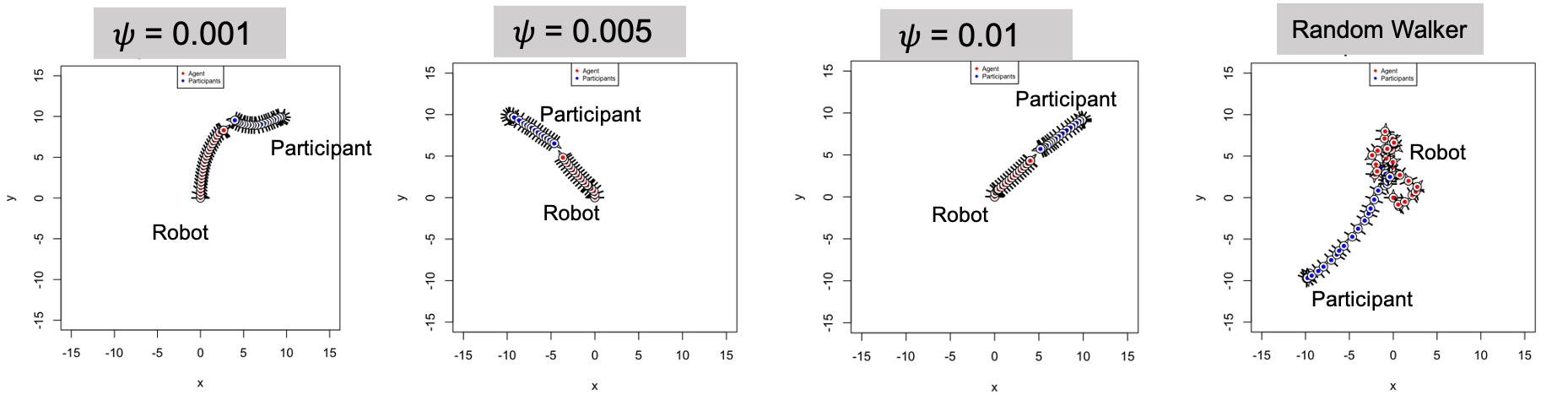}
		\caption{Example trajectories of a participant and a robot in a task where an object held by the robot is specified as a target}\label{fig:res_trj_app}
\end{figure*}

Figure \ref{fig:res_trj_app} shows an example of the observed interaction between the participant and the virtual robot when an object held by the robot was designated as the target in each experimental condition. As the trajectories in Fig. \ref{fig:res_trj_app} show, the robot moving based on the model also approached the approaching participant. In contrast, the participant’s trajectory is again longer for the random-walking robot, as shown by the rightmost trajectory in Fig. \ref{fig:res_trj_app}.

\begin{figure}[tb]
	\centering
		\includegraphics[width=\linewidth]{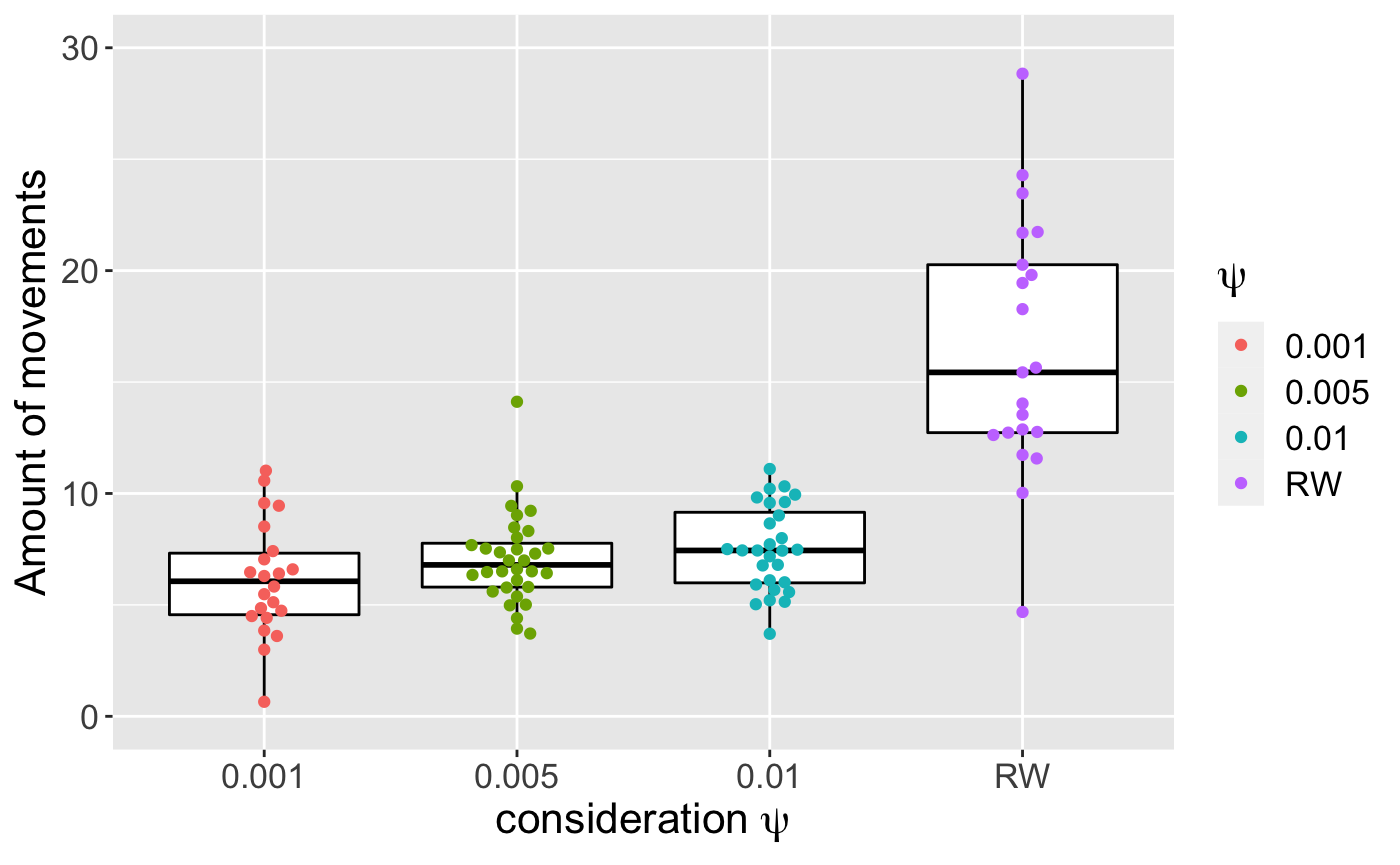}
		\caption{Comparison of participant movements when approaching the robot under each condition}\label{fig:res_mov}
\end{figure}

Figure \ref{fig:res_mov} shows the total amount of movement (distance traveled) when the participant goes toward the object held by the virtual robot. Participants again moved more in the RW condition ($M=23.29$, $SD=12.24$). On the other hand, there was little difference among results in the $\psi_{\rm Robot}=0.001$, $\psi_{\rm Robot}=0.005$, and $\psi_{\rm Robot}=0.01$ conditions ($M=6.15$, $SD=2.23$; $M=7.04$, $SD=1.73$, $M=7.51$, $SD=1.55$, respectively).
One-way between-group ANOVA results show significant differences in movement among the four conditions ($F(3,50)=21.13$, $f=1.13$, $p < 0.01$).
The results of multiple comparisons by HSD show that the RW condition had a significantly higher amount of movement than the other three conditions. The movement differences among the $\psi_{\rm Robot}=0.001$, $\psi_{\rm Robot}=0.005$ and $\psi_{\rm Robot}=0.01$ conditions were not significant for any of the combinations compared. In cases where the participant approached the robot, the robot based on the model approached the participant at any value of consideration. These results support H3. 

\subsection{Discussion}
We conducted experiments in a virtual environment to demonstrate the validity of the proposed model. As four experimental conditions, we compared participant interactions with a robot using consideration parameters $\psi_{\rm Robot}=0.001$, $\psi_{\rm Robot}$=0.005, and $\psi_{\rm Robot}=0.01$, as well as a randomly walking robot (RW). Two working hypotheses, H1, H2 and H3, were tested to validate the model.

H1 states that a robot with a low consideration parameter inhibits the movement of a participant toward a pole. Our analysis shows that the gap between the participant’s direction of movement and the pole direction was significantly larger with the robot consideration parameter $\psi_{\rm Robot}=0.001$ than with the other consideration parameters. This result supports H1. 

H2 posits that repeated interactions with robots having a lower consideration value $\psi_{\rm Robot}$ increase the tendency of individuals to avoid those robots. By using Bayesian statistics to estimate the trend of increased avoidance behavior with the number of movements within a trial, it was shown that under the condition of a low consideration value for the robot ($\psi_{\rm Robot}=0.001$), there is a 96.2\% probability that this increasing trend is positive. This result supports Hypothesis H2.

H3 says that when a participant approaches an object held by a virtual robot, the robot approaches the participant regardless of the consideration value. In cases where the participant approached the robot, the robot based on the model indeed approached the participant at all values of consideration. This result supports H3. 

All of the working hypotheses (H1, H2, H3) were derived from simulation results of the model, and the evidence supporting all three of them suggests that the proposed model is a valid tool for clarifying the effect of consideration in human–robot interaction.

By extending the model of a previous study, we were able to construct a model that explicitly incorporates social consideration for others. When a robot with a low consideration value approaches a human, it inhibits the human’s movement. On the other hand, when the consideration value is larger, the robot can approach the person without interfering with the person’s movement. In addition, the robot can also approach the person when the person first approaches it. The robot based on the proposed model estimates the internal state of others from their actions and adjusts its internal state accordingly. This adjustment of the internal state allows the generation of behavior reflecting the above observations.

The robot repeatedly approached humans without applying consideration, which prompted them to avoid it. As shown in Fig. \ref{fig:res_beta}, participants became more evasive toward robots with low consideration values as the number of tasks increased. On the other hand, there was no tendency for avoidance behavior to increase when the consideration value was higher. These results suggest that the model of consideration for others can effectively generate behaviors for communication robots in public situations. However, statistical analysis using a generalized mixed model is necessary to support this claim more conclusively.

How persistently a robot initiates communication with others depends on the situation: the robot’s role, the content of the dialogue, the ambiance of the environment, and the attributes of the people around it. For example, a guidance robot in a hotel or airport might need to set a higher value of consideration than a robot engaged in advertising in a shopping mall. The proposed model can easily adjust the robot’s behavior, according to the various circumstances, by setting the initial value of the robot’s internal state and the amount of consideration. This model describes the internal states and considerations of agents, including people. Therefore, it is also possible to compute the distribution of the initial parameter values of the internal states and consideration levels of the humans who pass by the location of the robot’s installation. Furthermore, the optimal robot parameters can be calculated in advance through simulation or adjusted through interaction. Future work will require field experiments using these methods.

\subsection{Limitations}
In this study’s experiments, a VR environment was used to observe the interactions between humans and virtual robots. While VR experiments offer the advantage of directly applying computational models and minimizing physical repercussions of proximity, they may also yield results that differ from real-world interactions. For instance, the motor noise emitted by physical robots can significantly affect the acceptance of their approach in real spaces \cite{joosse2021making}. Although experiments with virtual robots present the benefit of leveraging a computational model as is, it is still essential to conduct future experiments with actual robots to advance toward practical application of the model, ensuring its relevance and effectiveness in real-world scenarios. 

This study analyzed the differences in interactions based on varying levels of consideration. However, it did not attempt to determine the optimal value of consideration. What constitutes such an optimal level likely depends on the role the robot is playing. For instance, in roles where the robot’s purpose is advertising or issuing warnings, persistent engagement by the robot might be beneficial. Conversely, for roles involving reception or support tasks, setting a higher value of consideration may be preferable. Consequently, it is necessary to explore the optimal levels of consideration according to the actual roles undertaken by robots and the specific contexts in which they are placed. Addressing this issue will likely require that experiments be conducted in various real-world settings to evaluate the effectiveness of different levels of consideration.

\section{Conclusions}
In this study, we constructed a model of interaction between agents that have consideration for others and tested the model’s validity through experiments in a virtual environment. A virtual robot estimated participants’ communication desires from their movements and then performed approach behavior while adjusting its own communication desires. We defined the amount of adjustment to the other’s communication needs as a parameter of consideration and analyzed its effect on the robot’s interaction with the participant depending on the value assigned to this parameter. The results show that when the value of consideration was low, the robot inhibited the participant’s movement path and prompted the participant to avoid the robot. On the other hand, when the consideration value was higher, the participant’s movement was not inhibited. Consequently, the proposed model is a valid tool for clarifying the effect of giving consideration to others in human-robot interaction.

\backmatter

\bmhead{Funding}
This work was supported by JSPS KAKENHI Grant Number 22H04862

\section*{Declarations}
\subsection*{Competing Interests}
The authors declare that they have no competing interests. 

\subsection*{Consent to Participate}
Informed consent was obtained from all individual participants involved in the study. 
This study was approved by the Ethics Committee on Human Research, Shizuoka University.

\subsection*{Data  Availability}
The data that support the findings of this study are available from the corresponding author, T. S., upon reasonable request.







\bibliographystyle{sn-basic}
\bibliography{social-robotics}

\end{document}